\documentclass[10pt,twocolumn]{IEEEtran}%
\usepackage[linesnumbered,ruled]{algorithm2e}
\usepackage{algorithmic}
\usepackage{amsmath}
\usepackage{amssymb}
\usepackage{amsfonts}
\usepackage{comment}
\usepackage{fullpage}
\usepackage{graphicx}
\usepackage{subfigure}
\usepackage{times}
\usepackage{epstopdf}
\usepackage{placeins}
\usepackage{amsmath,bm,times}
\usepackage{cjhebrew}%
\providecommand{\U}[1]{\protect\rule{.1in}{.1in}}
\newcommand{\url}{}

\graphicspath{{./figures/}}
\begin{document}

\title{Auto-ML\ Deep Learning for Rashi Scripts OCR}
\author{Shahar Mahpod\thanks{{Faculty of Engineering, Bar Ilan University, Israel}{}.
mahpod.shahar@gmail.com.}, Yosi Keller\thanks{{Faculty of Engineering, Bar
Ilan University, Israel{}}{}. yosi.keller@gmail.com.}}
\date{}
\maketitle

\begin{abstract}
In this work we propose an OCR scheme for manuscripts printed in Rashi font
that is an ancient Hebrew font and corresponding dialect used in religious
Jewish literature, for more than 600 years. The proposed scheme utilizes a
convolution neural network (CNN) for visual inference and Long-Short Term
Memory (LSTM) to learn the Rashi scripts dialect. In particular, we derive an
AutoML scheme to optimize the CNN architecture, and a book-specific CNN
training to improve the OCR accuracy. The proposed scheme achieved an accuracy
of more than 99.8\% using a dataset of more than 3M annotated letters from the
Responsa Project dataset.

\end{abstract}

\section{Introduction}

\label{sec:Introduction}

The Optical Character Recognition (OCR) of printed media such as books and
newspapers is of major importance, as it enables digital access, archiving,
search and Natural Language Processing (NLP) based analysis of texts. A gamut
of digitization projects, such as the America's Historical Newspapers,
1690-1922\footnote{http://www.readex.com}, the California Digital Newspaper
Collection\footnote{http://cdnc.ucr.edu/cgi-bin/cdnc}, the Doria project, and
the digitization Project of Kindred Languages\footnote{http://www.doria.fi},
to name a few, were established. Some OCR projects deal with Hebrew language
digitization, such as the Historical Jewish Press
project\footnote{http://web.nli.org.il/sites/JPress/Hebrew/Pages/default.aspx}%
, and the Early Hebrew Newspapers
Project\footnote{http://jnul.huji.ac.il/dl/newspapers/index1024.html}, while
others, such as the Responsa Project\footnote{http://www.responsa.co.il},
Historical Dictionary Project\footnote{hebrew-treasures.huji.ac.il}, Otzar
HaHochma\footnote{http://www.otzar.org/wotzar} and
HebrewBooks\footnote{http://hebrewbooks.org/}, focus on religious Jewish
literature. It is common in such projects to utilize commercial OCR packages
such as ABBYY\footnote{https://www.abbyy.com/} or
OmniPage\footnote{http://www.nuance.com/for-individuals/by-product/omnipage/index.htm}%
.

The Responsa Project is one of the largest scale digitizations of Hebrew and
Jewish books. In particular, it specializes in Rashi script OCR that is an
ancient typeface of the Hebrew alphabet, based on a 15th century Sephardic
semi-cursive handwriting. It is has been used extensively in Jewish religious
literature and Judaeo-Spanish books for more than 600 years, and is in large
scale use nowadays. Rashi and Hebrew scripts letters are depicted in Table
\ref{tab:rashi-fonts}. We denote as \textit{characters} the different
manifestations (in different fonts) of the same underlying \textit{letters}.

Pattern Matching \cite{953958} and Feature Extraction \cite{6195349,1227699}
were applied to OCR, by comparing series of image descriptors encoding the
characters, such as SIFT \cite{6195349}, SURF
\cite{Bay:2008:SRF:1370312.1370556}, and PCA \cite{1227699} to name a few. The
recognition was formulated as a classification task using SVM
\cite{1227699,6195349}. OCR was one of the first applications of Convolutional
Neural Networks (CNNs), due to the seminal work by LeCun et al.
\cite{LeCun:1989:BAH:1351079.1351090}.

The recognition accuracy of OCR can be improved by applying NLP to the visual
analysis results. NLP was applied using dictionaries \cite{6751207,953957},
statistical algorithms such as HMM \cite{DBLP:conf/drr/2005}, Graph
optimization \cite{6247990} and LSTM \cite{4531750}. Contemporary CNN-based
NLP\ schemes often utilize word embeddings \cite{icml2014c2_santos14},
alongside RNN and LSTM layers. The work of Kim et al.
\cite{DBLP:journals/corr/KimJSR15}, is of particular interest as multiple
stacked LSTM layers were used to compute a hierarchy of embeddings, first
embedding the sequence of letters to represent words, and the sequence of
words to represent sentences.\begin{table}[tbh]
\centering
\par%
\begin{tabular}
[c]{|cc|cc|cc|cc|cc|}\hline
\multicolumn{2}{|c|}{} & \multicolumn{2}{|c|}{} & \multicolumn{2}{|c|}{} &
\multicolumn{2}{|c|}{} & \multicolumn{2}{|c|}{}\\
{\LARGE {\textcjheb{'}}} & \includegraphics[width=0.02\textwidth]{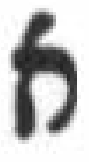} &
{\LARGE {\textcjheb{b}}} & \includegraphics[width=0.02\textwidth]{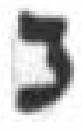} &
{\LARGE {\textcjheb{g}}} & \includegraphics[width=0.02\textwidth]{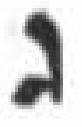} &
{\LARGE {\textcjheb{d}}} & \includegraphics[width=0.02\textwidth]{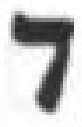} &
{\LARGE {\textcjheb{h}}} & \includegraphics[width=0.02\textwidth]{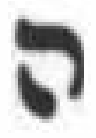}\\
\multicolumn{2}{|c|}{alf} & \multicolumn{2}{|c|}{bet} &
\multicolumn{2}{|c|}{gml} & \multicolumn{2}{|c|}{dlt} &
\multicolumn{2}{|c|}{hea}\\\hline
\multicolumn{2}{|c|}{} & \multicolumn{2}{|c|}{} & \multicolumn{2}{|c|}{} &
\multicolumn{2}{|c|}{} & \multicolumn{2}{|c|}{}\\
{\LARGE {\textcjheb{w}}} & \includegraphics[width=0.02\textwidth]{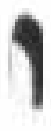} &
{\LARGE {\textcjheb{z}}} & \includegraphics[width=0.02\textwidth]{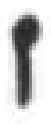} &
{\LARGE {\textcjheb{.h}}} & \includegraphics[width=0.02\textwidth]{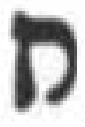} &
{\LARGE {\textcjheb{.t}}} & \includegraphics[width=0.02\textwidth]{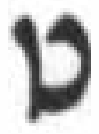} &
{\LARGE {\textcjheb{y}}} & \includegraphics[width=0.02\textwidth]{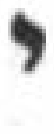}\\
\multicolumn{2}{|c|}{vav} & \multicolumn{2}{|c|}{zin} &
\multicolumn{2}{|c|}{het} & \multicolumn{2}{|c|}{tet} &
\multicolumn{2}{|c|}{yod}\\\hline
\multicolumn{2}{|c|}{} & \multicolumn{2}{|c|}{} & \multicolumn{2}{|c|}{} &
\multicolumn{2}{|c|}{} & \multicolumn{2}{|c|}{}\\
{\LARGE {\textcjheb{\<K>}}} &
\includegraphics[width=0.02\textwidth]{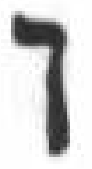} &
{\LARGE {\textcjheb{\<k|>}}} &
\includegraphics[width=0.02\textwidth]{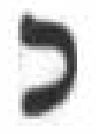} & {\LARGE {\textcjheb{l}}} &
\includegraphics[width=0.02\textwidth]{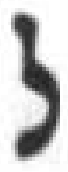} & {\LARGE {\textcjheb{\<M>}}}
& \includegraphics[width=0.02\textwidth]{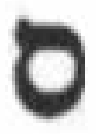} &
{\LARGE {\textcjheb{\<m|>}}} &
\includegraphics[width=0.02\textwidth]{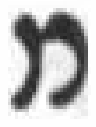}\\
\multicolumn{2}{|c|}{kaf-p} & \multicolumn{2}{|c|}{kaf} &
\multicolumn{2}{|c|}{lmd} & \multicolumn{2}{|c|}{mem-p} &
\multicolumn{2}{|c|}{mem}\\\hline
\multicolumn{2}{|c|}{} & \multicolumn{2}{|c|}{} & \multicolumn{2}{|c|}{} &
\multicolumn{2}{|c|}{} & \multicolumn{2}{|c|}{}\\
{\LARGE {\textcjheb{\<N>}}} &
\includegraphics[width=0.02\textwidth]{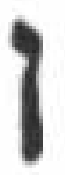} &
{\LARGE {\textcjheb{\<n|>}}} &
\includegraphics[width=0.02\textwidth]{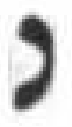} & {\LARGE {\textcjheb{s}}} &
\includegraphics[width=0.02\textwidth]{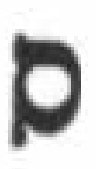} & {\LARGE {\textcjheb{`}}} &
\includegraphics[width=0.02\textwidth]{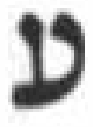} & {\LARGE {\textcjheb{\<P>}}}
& \includegraphics[width=0.02\textwidth]{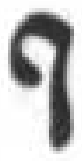}\\
\multicolumn{2}{|c|}{nun-p} & \multicolumn{2}{|c|}{nun} &
\multicolumn{2}{|c|}{smk} & \multicolumn{2}{|c|}{ain} &
\multicolumn{2}{|c|}{peh-p}\\\hline
\multicolumn{2}{|c|}{} & \multicolumn{2}{|c|}{} & \multicolumn{2}{|c|}{} &
\multicolumn{2}{|c|}{} & \multicolumn{2}{|c|}{}\\
{\LARGE {\textcjheb{\<p|>}}} &
\includegraphics[width=0.02\textwidth]{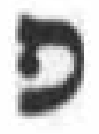} & {\LARGE {\textcjheb{\<.S>}
}} & \includegraphics[width=0.02\textwidth]{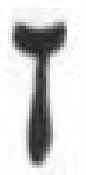} &
{\LARGE {\textcjheb{\<.s|>}}} &
\includegraphics[width=0.02\textwidth]{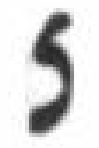} & {\LARGE {\textcjheb{q}}} &
\includegraphics[width=0.02\textwidth]{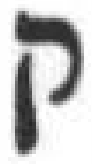} & {\LARGE {\textcjheb{r}}} &
\includegraphics[width=0.02\textwidth]{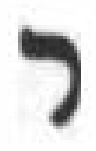}\\
\multicolumn{2}{|c|}{peh} & \multicolumn{2}{|c|}{zdk-p} &
\multicolumn{2}{|c|}{zdk} & \multicolumn{2}{|c|}{kuf} &
\multicolumn{2}{|c|}{rsh}\\\hline
\multicolumn{2}{|c|}{} & \multicolumn{2}{|c|}{} & \multicolumn{6}{|c|}{}\\
{\LARGE {\textcjheb{/s}}} & \includegraphics[width=0.02\textwidth]{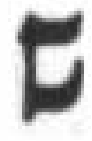} &
{\LARGE {\textcjheb{t} }} & \includegraphics[width=0.02\textwidth]{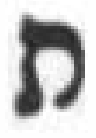} &
\multicolumn{6}{|c|}{}\\
\multicolumn{2}{|c|}{shn} & \multicolumn{2}{|c|}{tav} & \multicolumn{6}{|c|}{}%
\\\hline
\end{tabular}
\caption{Printed Hebrew characters and the corresponding Rashi characters. The
left character in each cell is the printed Hebrew character, while the right
character is the corresponding Rashi character.}%
\label{tab:rashi-fonts}%
\end{table}

\begin{figure}[tbh]
\centering
\includegraphics[width=0.5\textwidth]{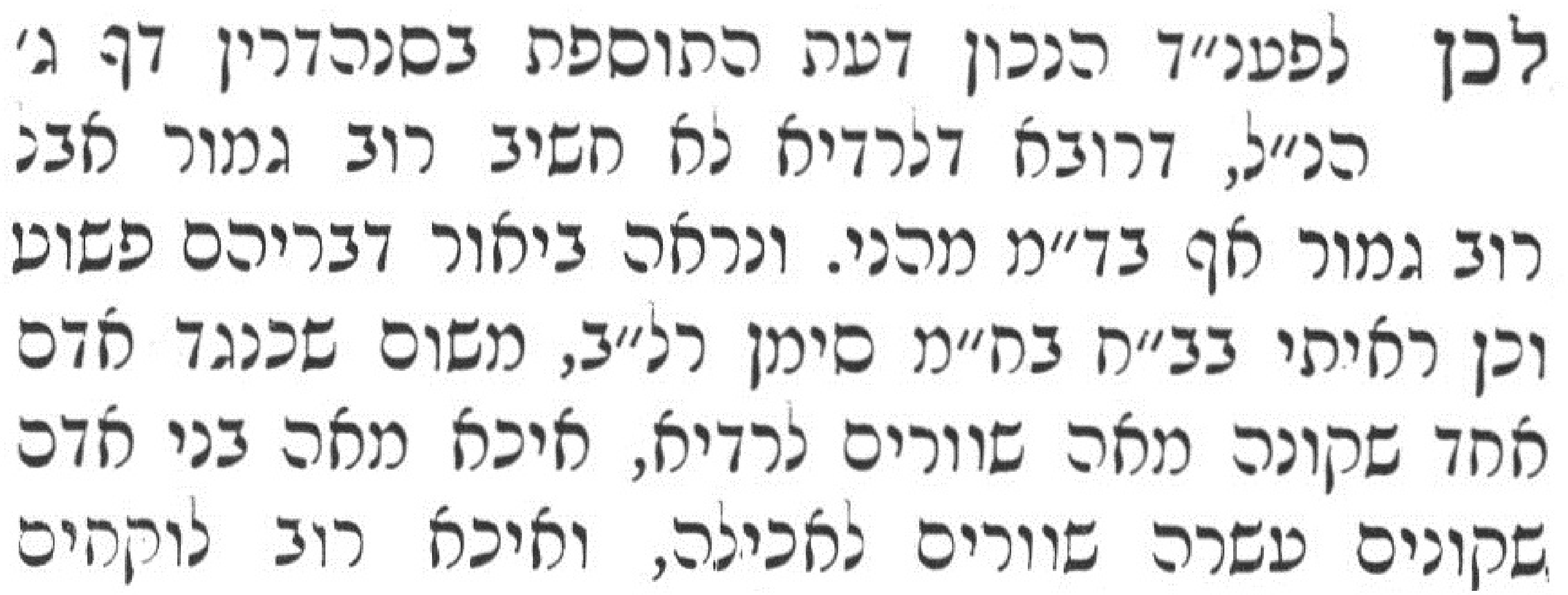} \newline%
\caption{An example of a Rashi script manuscript. Part of a manuscript printed
in Rashi script, taken from \textquotedblleft Shut Rabeynu Yosef
Mislutsk\textquotedblright\ manuscript, printed in Slutsk (Belarus), at the
first half of the 19 century.}%
\label{fig:new-book}%
\end{figure}

In this work we study the OCR of Rashi scripts that entails several
challenges. First, being an exotic font, there are no commercial or academic
OCR softwares for creating a training set. Second, it is common in some books
to utilize the Rashi font as a primary font, alongside regular printed Hebrew
fonts to emphasize the beginning of paragraphs, as shown in Fig.
\ref{fig:new-book}, or mark citations from older books, such as the Bible or
the Talmud.

\begin{figure}[tbh]
\centering
\par%
\begin{tabular}
[c]{ccccc}%
\includegraphics[width=0.03\textwidth]{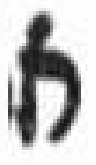} &
\includegraphics[width=0.03\textwidth]{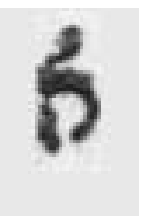} &
\includegraphics[width=0.03\textwidth]{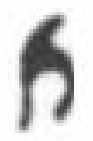} &
\includegraphics[width=0.03\textwidth]{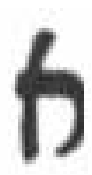} &
\includegraphics[width=0.03\textwidth]{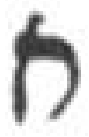}\\
\multicolumn{5}{c}{\textbf{a}}\\
\includegraphics[width=0.03\textwidth]{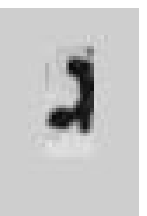} &
\includegraphics[width=0.03\textwidth]{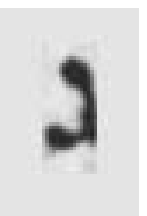} &
\includegraphics[width=0.03\textwidth]{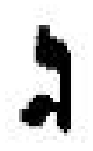} &
\includegraphics[width=0.03\textwidth]{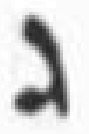} &
\includegraphics[width=0.03\textwidth]{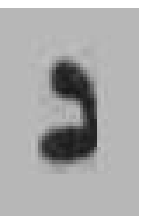}\\
\multicolumn{5}{c}{\textbf{b}}%
\end{tabular}
\par
. \caption{The variability in Rashi script letters appearing in different
books. (a) Samples of the letter \textcjheb{'} (alf). (b) Samples of the
letter \textcjheb{g} (gml).}%
\label{fig:various-sample}%
\end{figure}

\begin{figure}[tbh]
\centering
\par%
\begin{tabular}
[c]{ccccc}%
\includegraphics[width=0.03\textwidth]{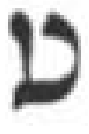} &
\includegraphics[width=0.03\textwidth]{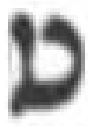} &
\includegraphics[width=0.03\textwidth]{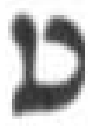} &
\includegraphics[width=0.03\textwidth]{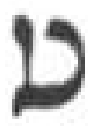} &
\includegraphics[width=0.03\textwidth]{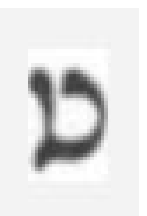}\\
\multicolumn{5}{c}{\textbf{a}}\\
\includegraphics[width=0.03\textwidth]{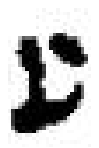} &
\includegraphics[width=0.03\textwidth]{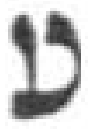} &
\includegraphics[width=0.03\textwidth]{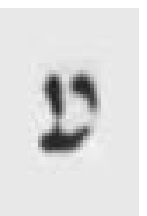} &
\includegraphics[width=0.03\textwidth]{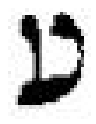} &
\includegraphics[width=0.03\textwidth]{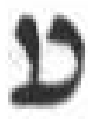}\\
\multicolumn{5}{c}{\textbf{b}}%
\end{tabular}
\caption{The visual similarity of different letters. (a) The letter
\textcjheb{.t} (tet) is visually similar to (b) \textcjheb{`} (ain).}%
\label{fig:rashi-fonts-ain-tet}%
\end{figure}

Third, the Rashi script books were printed by different printing houses, over
more than 500 years, resulting in significant variations of the script, as
depicted in Fig. \ref{fig:various-sample}. For instance, the pair of different
characters \{'ain','tet'\} , \{'dlt','rsh'\} \{'nun-p','zdk-p'\} are similar,
while the same letters printed by different printing houses might look
different, as depicted in Fig. \ref{fig:rashi-fonts-ain-tet}.


\begin{figure}[tbh]
\centering
\includegraphics[width=0.5\textwidth]{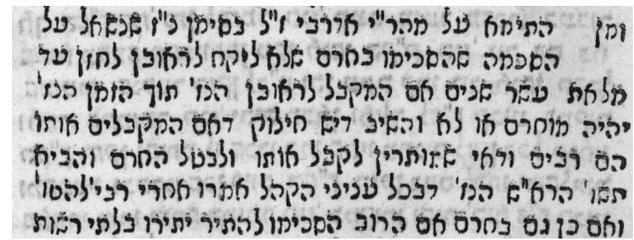} \caption{An
example of a corrupted manuscript page scanned from the book \textquotedblleft
Pney Aharon\textquotedblright, printed in Saloniki, Greece, at the first half
of the 18th century.}%
\label{fig:old-book}%
\end{figure}

\begin{figure}[tbh]
\centering
\par%
\begin{tabular}
[c]{ccccc}%
\includegraphics[width=0.03\textwidth]{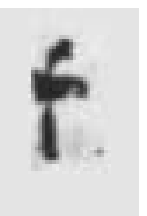} &
\includegraphics[width=0.03\textwidth]{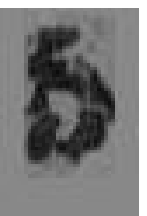} &
\includegraphics[width=0.03\textwidth]{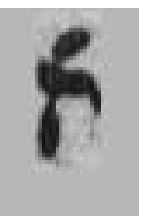} &
\includegraphics[width=0.03\textwidth]{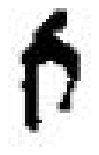} &
\includegraphics[width=0.03\textwidth]{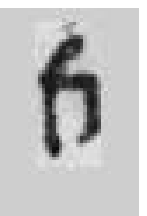}\\
\includegraphics[width=0.03\textwidth]{./1490-074.eps} &
\includegraphics[width=0.03\textwidth]{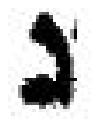} &
\includegraphics[width=0.03\textwidth]{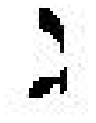} &
\includegraphics[width=0.03\textwidth]{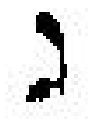} &
\includegraphics[width=0.03\textwidth]{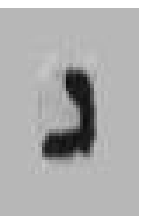}
\end{tabular}
\caption{Examples of corrupted and distorted Rashi script letters scanned in
different books. The top line shows the samples of the letter \textcjheb{'}
(alf), while the bottom line depicts the letter \textcjheb{g} (gml). }%
\label{fig:rashi-fonts-corrupt}%
\end{figure}

Other issues relate to handling scanning effects such as impurities, skewed
text lines, and page folds as depicted in Figs. \ref{fig:old-book} and
\ref{fig:rashi-fonts-corrupt}.

\begin{figure}[tbh]
\centering
\par%
\begin{tabular}
[c]{cc}%
\includegraphics[width=0.04\textwidth]{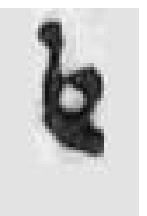} &
\includegraphics[width=0.04\textwidth]{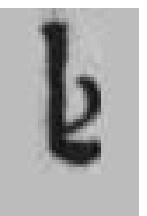}
\end{tabular}
\caption{Special and rare symbols in Rashi scripts. This special symbol
combines the letters \textcjheb{'} (alf) and \textcjheb{l} (lmd).}%
\label{fig:alf-lmd}%
\end{figure}

Some ancient Jewish books are fully or partially written in Aramaic language,
contain special symbols for acronyms, and a rare composition of \textcjheb{'}
(alf) and \textcjheb{l} (lmd) as in Fig. \ref{fig:alf-lmd}. Implying that the
OCR schemes can not apply standard dictionaries and spelling correction
schemes, that have to be learnt as part of the OCR scheme. As the Rashi script
consists of disconnected characters, the agnostic detection of isolated
characters, that is the detection of the bounding box of each character
without detecting the character's class, can be easily
implemented.

In this work we propose a Deep Learning based OCR scheme for learning both the
visual representation of the Rashi script, and the spelling of the
corresponding manuscripts. This is a particular example of scripts in an
exotic language, where both the font and spelling are apriori unknown. The
proposed scheme is fully data driven and does not utilize predefined
dictionaries. We derive a unified Deep network that combines a CNN to visually
classify the characters, with a LSTM to learn the corresponding spelling, and
thus improve the OCR accuracy.

In particular, we propose the following contributions:

\textbf{First,} we derive a Deep network consisting of a CNN trained over 3M
Rashi script samples, and a LSTM layer trained over words, that is shown to
yield highly accurate OCR results.

\textbf{Second}, we derive an Auto-ML scheme based on Genetic Algorithms to
optimally design the structure of the CNN. This results in an improved OCR
accuracy achieved by a CNN three-fold smaller.

\textbf{Last}, we propose an active learning approach for refining the net
accuracy when applied to a particular book, by refining the CNN model using a
small set of the test characters.

The rest of this paper is organized as follows: we start by reviewing previous
works on deep learning based OCR in Section \ref{sec:Related Work}. The
proposed approach is introduced in Section \ref{sec:our approach}, while the
experimental validation and comparison between different schemes is presented
in Section \ref{sec:exp}. Concluding remarks and future work are discussed in
Section \ref{sec:conclusions}.

\section{Related Work}

\label{sec:Related Work}

Optical Character Recognition (OCR) is a common task in computer vision, used
in a gamut of applications and languages such as, English \cite{Jaderberg14c}%
\cite{Jaderberg14d}, Chinese \cite{7783877} and Japanese \cite{6021648}. The
OCR of exotic languages such as the Rashi script is less applicative and common.

Similar to our work, Deep Learning (DL) was applied to exotic languages. A
LeNet-based CNN\ \cite{726791} was applied by Rakesh et al.
\cite{journals/corr/AchantaH15} for the OCR of the Telugu script consisting of
$\sim460$ symbols that were encoded by binary image patches. A similar
architecture was applied by Kim et al. \cite{Kim:2015:HHR:2738735.2738757} for
the OCR of the handwritten Hangul script, while using the MSE as a loss
function. A weighted MSE was proposed to handle the imbalanced training set
during training.

A deep belief network (DBN), with three fully connected layers, was proposed
by Sazal et al. \cite{6777907} for the Bangla handwritten OCR dataset
consisting of ten numerals and 50 characters. The layers were initialized
using a restricted Boltzmann machine (RBM), and a softmax loss was used to
classify the characters. A similar approach was applied by Ma et al.
\cite{6981039} for the OCR of a Tibetan script consisting of 562 characters
and applied a three-layers DBM initialized using a RBM. SIFT local image
descriptors were applied by Sushma and Veena \cite{7779380} to encode and
detect the characters of the Kannada language that uses 49 phonemic
characters, where different characters can be composed to encode a single
symbol. A Hidden Markov Model (HMM) was applied to improve the decoding of the
language symbols by training the HMM using texts.

Deeper CNNs were used by Zhong et al. \cite{DBLP:journals/corr/ZhongJX15} for
the Handwritten Chinese Character Recognition (HCCR) competition, by
considering AlexNet \cite{NIPS2012_4824} and GoogLeNet
\cite{DBLP:journals/corr/SzegedyLJSRAEVR14}, where the images of the
characters, their image gradients, HOG and Gabor descriptors were used as
features. Stacked RNNs, the first for script identification and the second for
recognition were applied by Mathew et al. \cite{7490115} to the OCR of
multilingual Indic Scripts, such as the Kannada, Bangla, Telugu and other
languages. Both networks consisted of three hidden layers and a LSTM layer.
Fuzzy logic and spatial image statistics were used by Gur and Zelavsky
\cite{6424419} to recognize distorted letters by using the combination of
letter statistics and correlation coefficients.

OCR is a particular example of the Structured Image Classification problem
where the classification problem is hierarchial. Such that the lower layers
analyze the visual data, and the succeeding layers analyze the inner
(semantic) structure of the data. In the proposed OCR scheme, the lower
inference layer is implemented by a CNN that classifies the visual
manifestations of the characters, and the semantic inference sub-network
learns the particular dialect used in Rashi scripts using LSTM. A Structured
Image Classification problem was studied by Goldman and Goldberger
\cite{DBLP:journals/corr/GoldmanG17} in the context of structured object
detection, where rows of products are detected in images by a CNN-based object
detection scheme. The order (structure) of the detected objects is learnt by
embedding the indexes of triplets of neighboring objects by an embedding
layer, and concatenating the structure embeddings with the Fully Connected
(FC) layer of the object detector. The resulting vector is used to infer the
true label of the input image. In contrast, the proposed scheme processes a
\textit{sequence} of object images simultaneously, and the sequential data is
encoded by a LSTM layer. Bar et al. detect compression fractures in CT scans
\cite{zebra} of the human spine. The spine that is a linear structure, was
first segmented, and its corresponding patches were binary classified using a
CNN. Recurrent Neural Network (RNN) was used to predict the existence of a
fracture. Our proposed scheme infers multiple labels both by the CNN layers
(characters) and LSTM\ (words).

\section{OCR of Rashi scripts using Deep Learning}

\label{sec:our approach}

\begin{figure}[tbh]
\centering	  
\includegraphics[width=0.45\textwidth]{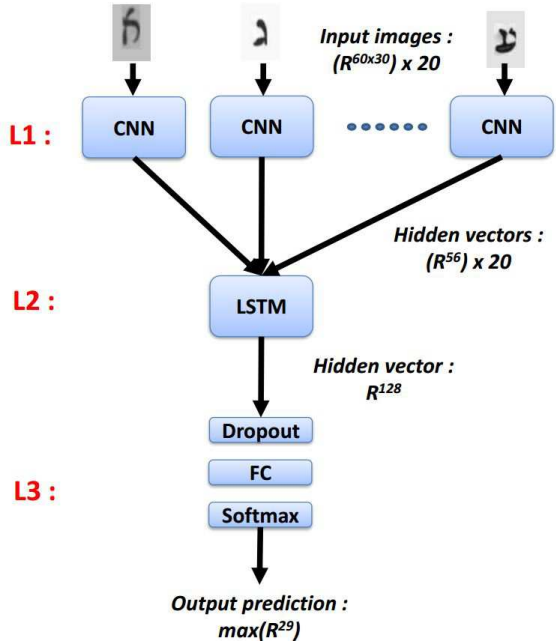} \caption{The proposed OCR
scheme, consisting of the CNN (L1) that classifies the characters' images and
outputs prediction probabilities per class, while the LSTM (L2) agglomerates
the probabilistic characters detections into words and sentences.}%
\label{fig:LSTM}%
\end{figure}

We study the OCR of Rashi scripts where the locations of the characters are
initially detected. Hence, the OCR is formulated as a classification problem,
where given a sequence of images of characters $\{\phi_{1},...,\phi_{J}\}$, we
aim to classify $C(\phi_{j})\in\left\{  c_{1},...,c_{L}\right\}  $, that are
the latent character labels. We utilize a Deep Learning scheme consisting of
two stacked sub-networks. The initial one, is a CNN\ that computes the
detection probabilities of the characters based on the input images, while the
succeeding sub-network utilizes a LSTM layer to improve the decoding accuracy,
by learning a data-driven vocabulary. An overall view of the proposed scheme
is depicted in Fig. \ref{fig:LSTM}. 

We studied several CNN architectures for classifying the images of the
detected characters. The first denoted Simple NN (SNN) is based on the LeNet5
architecture \cite{726791} detailed in Table \ref{table:Simple net}%
.\begin{table}[tbh]
\centering%
\begin{tabular}
[c]{|c|c|c|c|}\hline
\textbf{Layer} & \textbf{Type} & \textbf{Parameters} & \textbf{Stride}%
\\\hline\hline
\textbf{L01} & \multicolumn{1}{|l|}{conv} & [5 5 1]x20 & \\\hline
& \multicolumn{1}{|l|}{max pool} & [2 2] & [2]\\\hline
\textbf{L02} & \multicolumn{1}{|l|}{conv} & [5 5 20]x50 & \\\hline
& \multicolumn{1}{|l|}{max pool} & [2 2] & [2]\\\hline
\textbf{L03} & \multicolumn{1}{|l|}{conv} & [4 4 50]x500 & \\\hline
& \multicolumn{1}{|l|}{max pool} & [2 1] & [4 1]\\\hline
\textbf{L04} & \multicolumn{1}{|l|}{conv} & [2 1 50]x500 & \\\hline
& \multicolumn{1}{|l|}{relu} &  & \\\hline
\textbf{L05} & \multicolumn{1}{|l|}{conv} & [1 1 500]x56 & \\\hline
& \multicolumn{1}{|l|}{softmax} &  & \\\hline
\end{tabular}
\caption{The CNN used for visual classification of the characters, based on
the LeNet5 CNN \cite{726791}. }%
\label{table:Simple net}%
\end{table}

Following the work of Zhong et al.\cite{DBLP:journals/corr/ZhongJX15}, we also
applied a deeper CNN\ based on AlexNet \cite{NIPS2012_4824}, detailed in Table
\ref{table:AlexNet}. \begin{table}[tbh]
\centering%
\begin{tabular}
[c]{|c|c|c|c|}\hline
\textbf{Layer} & \textbf{Type} & \textbf{Parameters} & \textbf{Stride}%
\\\hline\hline
\textbf{L01} & conv & [11 11 1]x96 & [2 1]\\\hline
& bnorm &  & \\\hline
& relu &  & \\\hline
& max pool & [3 3] & [2]\\\hline
\textbf{L02} & conv & [5 5 48]x256 & \\\hline
& bnorm &  & \\\hline
& relu &  & \\\hline
& max pool & [3 3] & \\\hline
\textbf{L03} & conv & [3 3 256]x384 & \\\hline
& bnorm &  & \\\hline
& relu &  & \\\hline
\textbf{L04} & conv & [3 3 384]x192 & \\\hline
& bnorm &  & \\\hline
& relu &  & \\\hline
\textbf{L05} & conv & [3 3 192]x256 & \\\hline
& bnorm &  & \\\hline
& relu &  & \\\hline
& max pool & [3 3] & \\\hline
\textbf{L06} & conv & [6 6 256]x4096 & \\\hline
& bnorm &  & \\\hline
& relu &  & \\\hline
\textbf{L07} & conv & [1 1 4096]x4096 & \\\hline
& bnorm &  & \\\hline
& relu &  & \\\hline
\textbf{L08} & conv & [1 1 4096]x56 & \\\hline
& softmax &  & \\\hline
\end{tabular}
\caption{The CNN used for visual classification of the characters, based on
the AlexNet architecture \cite{NIPS2012_4824}.}%
\label{table:AlexNet}%
\end{table}

The Spatial Transformer layer \cite{NIPS2015_5854}\cite{DBLP:journals/corr/ShiWLYB16} was shown to
improve OCR accuracy, by geometrically rectifying the input image by
estimating the underlying affine transformation. Thus, we added the Spatial
Transformer layer to the MNIST CNN, and denote the resulting CNN Spatial
Transformer Net (STN), as reported in Table \ref{table:Spatial net}%
.\begin{table}[tbh]
\centering%
\begin{tabular}
[c]{|c|c|c|c|}\hline
\textbf{Layer} & \textbf{Type} & \textbf{Parameters} & \textbf{Stride}%
\\\hline\hline
& max pool & [2 2] & [2 1]\\\hline
\textbf{L01} & conv & [5 5 1]x20 & \\\hline
& relu &  & \\\hline
& max pool & [2 2] & [2]\\\hline
\textbf{L02} & conv & [5 5 20]x20 & \\\hline
& relu &  & \\\hline
\textbf{L03} & conv & [9 9 20]x50 & \\\hline
& relu &  & \\\hline
\textbf{L04} & conv & [1 1 50]x6 & \\\hline
& grid &  & \\\hline
& sampler &  & \\\hline
& max pool & [2 2] & \\\hline
\textbf{L05} & conv & [7 7 1]x32 & \\\hline
& relu &  & \\\hline
& max pool & [2 2] & [2]\\\hline
\textbf{L06} & conv & [7 7 1]x32 & \\\hline
& relu &  & \\\hline
& max pool & [2 2] & [2]\\\hline
\textbf{L07} & conv & [4 4 32]x48 & \\\hline
& relu &  & \\\hline
& max pool & [2 2] & [2]\\\hline
\textbf{L08} & conv & [4 4 48]x256 & \\\hline
& relu &  & \\\hline
\textbf{L09} & conv & [1 1 256]x56 & \\\hline
& softmax &  & \\\hline
\end{tabular}
\caption{The CNN used for visual classification of the characters based on the
MNIST CNN and a Spatial Transformer layer \cite{NIPS2015_5854}. }%
\label{table:Spatial net}%
\end{table}

Thus, the output of the CNN is the classification probabilities, $\left\{
p_{j,l}\right\}  _{1,1}^{J,L}$ of the input sequence of character images
$\left\{  \phi_{j}\right\}  _{1}^{J}$, such that $p_{j,l}$ is the
\textit{estimated} probability of an image $\phi_{j}\in\left\{  \phi
_{j}\right\}  _{1}^{J}$ to be related to the class (letter) $c_{l}\in\left\{
c_{l}\right\}  _{1}^{L}$. As a script might contain both Rashi and printed
Hebrew fonts, we set $L=56$, consisting of \ 27 characters of Rashi script, 27
characters of Printed Hebrew script, one for a special character 'a-l' and a
single space character.

The scripts are written in a particular Hebrew dialect that can be utilized to
better infer the estimated symbols, as being part of words and sentences. For
that we apply Long-short term memory (LSTM)
\cite{doi:10.1162/neco.1997.9.8.1735} that allows to encode and utilize the
inherent sequential information. The LSTM layer is denotes as $L2$ in Fig.
\ref{fig:LSTM} and consists of 20 memory units, where the input to each LSTM
module is $\left\{  p_{j,l}\right\}  _{1}^{L}\in R^{56}$ , (20 vectors in
total), and the output in $R^{128}$ encodes the sequence of CNN outputs
$\left\{  p_{j,l}\right\}  _{1,1}^{J,L}$.

\subsection{AutoML CNN architecture refinement using a genetic algorithm}

\label{subsec:automl}

We applied AutoML to refine the CNNs proposed in Section
\ref{sec:our approach} by optimizing the classification accuracy with respect
to the CNN architecture, that is given by the parameters of its layers. In
particular, we consider a CNN similar to LeNet5 \cite{726791} consisting of
sequential pairs of convolution and activation layers that are jointly
optimized. Thus, we optimize the parameters of the convolution layers: the
support of the filter, the number of filters (output dimensionality), the
stride in both image axes, the use of batch normalization, and the use of a
dropout layer and its probability. The activation layer is given by its type
(ReLU, Avg-Pool, Max-Pool), where the pooling layers are given by their
pooling support and stride.

AutoML is a nonlinear optimization over a heterogeneous set of discrete
(filters support, stride etc.) categorical (activation type) and continuous
(dropout rate) parameters. For that we apply a Genetic Algorithm (GA)
\cite{Mitchell:1998:IGA:522098} by first (generation $g=0$) drawing $N=120$
random CNNs $\left\{  CNN_{i}^{g=0}\right\}  $ similar to LeNet5, each
consisting of $l$ convolution+activation layers, followed by a fixed FC layer.
Each such CNN is trained for $M=10$ epochs using $\sim5\%$ of the training
data. The $K_{1}$ CNNs having the highest validation accuracy are denoted as
the \textit{Elite} set, while $P=40\%$ of the remaining CNNs with the highest
validation accuracy are denoted as \textit{Parents}.

The next generation of CNNs $\left\{  CNN_{i}^{g+1}\right\}  $ is derived by
computing \textit{Children}, \textit{Mutations,}and\textit{ Random }CNNs:
\textit{Children} are computed by randomly picking pairs of \textit{Parents
}CNNs and splitting each of them randomly, and connecting the resulting
sub-CNNs, that are of varying lengths $\left\{  l_{n}\right\}  _{1}^{N}$.
\textit{Mutations} are created by picking a random parent and randomly
changing a single CNN parameter. Additional \textit{Random }CNNs are drawn as
in the initialization phase. Thus, $\left\{  CNN_{i}^{g+1}\right\}  $ is
initialized by propagating the \textit{Elite} set to $\left\{  CNN_{i}%
^{g+1}\right\}  $, and adding the $K_{2}$ CNNs in $\left\{
\mathit{Children\cup Mutations\cup Random}\right\}  $ having the highest
validation accuracy%
\begin{multline}
\left\{  CNN_{i}^{g+1}\right\}  =\\
\mathit{Elite\cup}\max_{K_{2}}\left\{  \mathit{Children\cup Mutations\cup
Random}\right\}  .
\end{multline}
\bigskip

This scheme is summarized in Algorithm \ref{alg_basic}, and was applied to
generate multiple consecutive generations. It is initialized by running
Algorithm \ref{alg_basic} for $l=\{3,5,7,9\}$ and selecting the resulting CNNs
having the highest validation accuracy. Algorithm \ref{alg_basic} is then
applied to this set of CNNs, and the resulting CNN is shown in Table
\ref{table:GANNNet10}.%

\begin{algorithm}[tbh]%
%

\begin{algorithmic}[1]%

\label{alg:ADL}\label{alg_basic}\caption{AutoML for CNN optimization using a Genetic Algorithm}\SetKwData{Left}{left}

\STATE\STATE\textbf{Initialization:} Generate $N$ random CNNs $\left\{
CNN_{i}^{g=0}\right\}  $ having $l$ convolution+activation layers.

\FOR{$g<G$}

\STATE Train $\left\{  CNN_{i}^{g}\right\}  $ for $M$ epochs

\STATE Apply $\left\{  CNN_{i}^{g}\right\}  $ on the validation set $V$

\STATE\textit{Elite=}$\max_{K_{1}}\left\{  CNN_{i}^{g}\left(  V\right)
\right\}  $

\STATE Compute the \textit{Children}, \textit{Mutations, }and \textit{Random
}sets of CNNs based on $V.$

\STATE$\left\{  CNN_{i}^{g+1}\right\}  =$\textit{Elite}$\cup$

$\max_{K_{2}}\left\{  \mathit{Children\cup Mutations\cup Random}\right\}  $

\ENDFOR%

\end{algorithmic}%
%

\end{algorithm}%
\begin{table}[tbh]
\centering%
\begin{tabular}
[c]{|c|c|c|c|}\hline
\textbf{Layer} & \textbf{Type} & \textbf{Parameters} & \textbf{Stride}%
\\\hline\hline
\textbf{L01} & \multicolumn{1}{|l|}{conv} & [7 7 1]x19 & [1 2]\\\hline
& \multicolumn{1}{|l|}{dropout} &  & \\\hline
& \multicolumn{1}{|l|}{avg pool} & [4 2] & \\\hline
\textbf{L02} & \multicolumn{1}{|l|}{conv} & [7 1 19]x83 & \\\hline
& \multicolumn{1}{|l|}{max pool} & [4 3] & [3 1]\\\hline
\textbf{L03} & \multicolumn{1}{|l|}{conv} & [14 9 83]x987 & [14 9]\\\hline
& \multicolumn{1}{|l|}{relu} &  & \\\hline
\textbf{L04} & \multicolumn{1}{|l|}{conv} & [1 1 987]x56 & \\\hline
& \multicolumn{1}{|l|}{softmax} &  & \\\hline
\end{tabular}
\caption{The OCR CNN (L1 in Fig. \ref{fig:LSTM}) computed by applying the
AutoML scheme.}%
\label{table:GANNNet10}%
\end{table}

\subsection{Book specific OCR-CNN refinement}
\label{subsec:Refine}

A particular attribute of the Rashi scriptures is that their Hebrew dialect
was and still is a sacred religious dialect and was not in daily use, as the
scholars using it lived in non-Hebrew speaking countries:\ Europe, North
Africa, etc. Thus, it did not undergo significant changes in 600 years, and
over different geographical regions. Hence, most Rashi manuscripts share a
similar dialect and vocabulary, but might differ in the graphical printing
attributes of the Rashi script, when printed by different printing houses, as
depicted in Fig. \ref{fig:various-sample}.

In order to utilize the dialect invariance, in contrast to the varying
graphical manifestation of the Rashi script letters, we propose to refine the
CNN \textit{per script} while freezing the LSTM layer, and use the
\textit{test} classifications as a training set for refining the CNN. For
that, given a test manuscript, we apply the CNN+LSTM as in Section
\ref{sec:our approach}, to $\sim5\%$ of the new script, and utilize the OCR
output as a training set for refining the CNN. The resulting CNN is applied to
the rest of the manuscript.

\section{Experimental Results}

\label{sec:exp}

\begin{figure}[tbh]
\centering
\par%
\begin{tabular}
[c]{c}%
\includegraphics[width=0.5\textwidth]{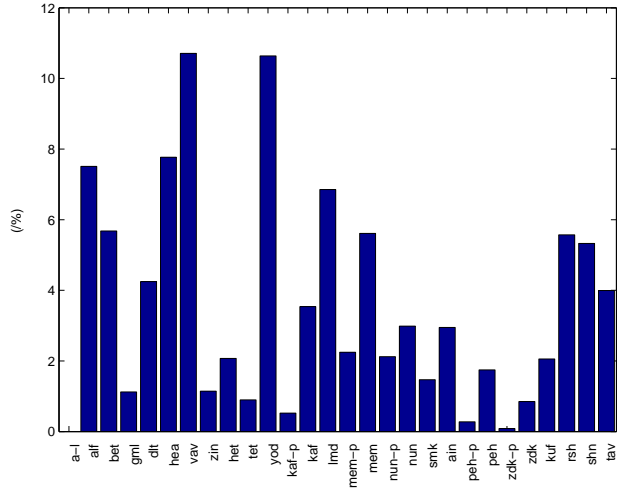}\\
\end{tabular}
\caption{The distribution of letters in the Responsa project dataset for both
Printed and Rashi scripts.}%
\label{fig:app-stat}%
\end{figure}

In this section we experimentally verify the validity and accuracy of the
proposed scheme by applying it to a large dataset provided by the Responsa
project consisting of $\sim$2500 pages extracted from 170 different books. The
dataset contains 5.5M annotated letters given their positions and labels.
There are on average 3400 letters per page, where some of the manuscripts
contain several pages, while others more then 20 pages. The images of the
letters are of size of 60x30 pixels, and their distribution is depicted in
Fig. \ref{fig:app-stat}.

We divided the Responsa project dataset into three sets $S^{r}$, $S^{v}$ and
$S^{t}$ consisting of 3.3M, 352K, and 2M letters, respectively. $S^{r}$ is the
training set consisting of 140 books and more than 1000 pages. $S^{v}$ was the
validation set (114 pages from 15 books), and $S^{t}$ was the testing set
collected from other 15 books. The test set $S^{t}$ was divided into two
subsets $S^{t}=S_{1}^{t}\cup S_{2}^{t}$, where $S_{1}^{t}$ consists of the
first 40K letters in each book, while the set $S_{2}^{t}$ comprises of the
remaining letters in a manuscript, 1.35M letters overall. $S_{1}^{t}$ was used
for the book-specific CNN\ refinement scheme introduced in Section
\ref{subsec:Refine}, while $S_{2}^{t}$ was used as a test set.

We applied the CNNs detailed in Section \ref{sec:our approach} and the CNN
computed using the AutoML scheme introduced in Section \ref{subsec:automl},
for which we used $2\%$ of the samples in the set $S^{r}$ and $10\%$ of
$S^{v}$, 75K and 35K letters, respectively. The AutoML refinement was applied
using seven different initial 'families', each was trained for $G=10$
generations and $120$ mutations each. The number of convolution layers, the
activations succeeding each convolution layer (ReLU, Max-Pool or Avg-Pool),
and the addition of batch normalization or dropout layers are randomly drawn.
Similarly, the layers' parameters (filters sizes, max pooling size, stride
value and dropout percentage) are also randomly drawn. The families differ by
the maximal number of initial layers.

\begin{figure}[tbh]
\centering
\includegraphics[width=0.5\textwidth]{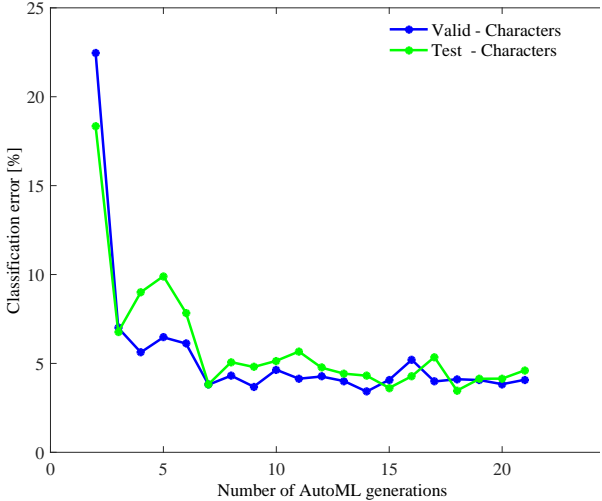}
\caption{The classification accuracy of the AutoML optimized CNN with respect
to the Genetic Algorithm's refinement iterations.}%
\label{fig:GANN-generations}%
\end{figure}

Other than the first generation, all of the mutations were chosen randomly,
the succeeding generations were composed of $K_{1}=5$ Elite group, 70
Cross-Over children and 25 mutations of the parents. The group of parents was
collected from the $K_{2}=40$ most accurate CNNs (including the Elite group)
computed in the previous generation. We also added additional 20 random
mutations at each generation to increase the variability. The  Elite group from the last generation of each family is gathered 
into a single set of CNNs and refined for $G=10$
generations and $120$ mutations. Figure \ref{fig:GANN-generations} depicts the
average classification error reduction with respect to the number of training
AutoML generations.

\subsection{OCR results using AutoML CNN}

\label{subsec:CNN Results}

\begin{figure}[tbh]
\centering
\includegraphics[width=0.5\textwidth]{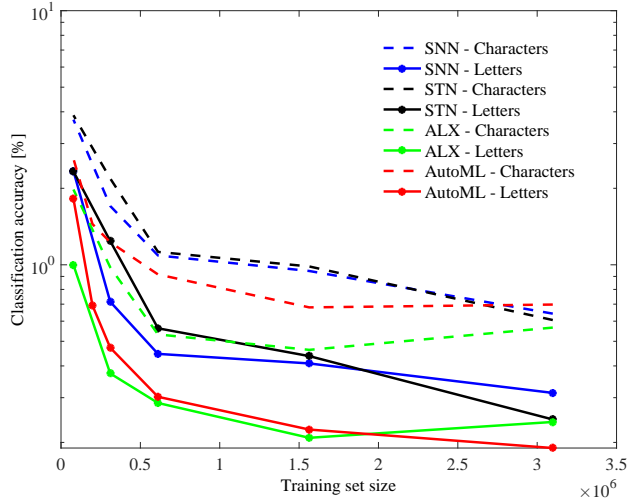}
\caption{The Characters and Letters classification accuracy of the proposed
CNNs, with respect o the training set size. STN - Spatial transform net. ALX -
AlexNet. AutoML - optimized by a Genetic Algorithm. }%
\label{fig:set-size-comparison}%
\end{figure}

We start by evaluating the proposed OCR scheme accuracy of the four CNNs
introduced in Section \ref{sec:our approach}, \textit{without} the additional
accuracy of the LSTM layer. The accuracy of the overall scheme (CNN+LSTM) is
discussed in Section \ref{subsec:LSTM Results}. Figure
\ref{fig:set-size-comparison} reports the classifications accuracy of the CNNs
with respect to the size of training set $S^{r}$. For each CNN we consider the
accuracy of detecting the 56 graphical symbols denoted as \textit{Characters},
and the accuracy of detecting the set of 29 \textit{phonetic} equivalents as
\textit{Letters}, where each \textit{Letter }(other than `space' and
punctuation marks) can\textit{ }be mapped to two \textit{Characters}, in Rashi
and Printed fonts. The accuracy results are also reported in Table
\ref{table:cnn summary}, where it follows that the SNN (based on the LeNet5
CNN) is inferior to the deeper AlexNet-based ALX and STN CNNs. The Spatial
Transformer layer does not improve the accuracy significantly compared to
AlexNet, and we attribute that to the deskewing of the OCR input images
applied in the preprocessing phase, such that the additional rectification of
the Spatial Transformer layer is insignificant.

The accuracy gap between the \textit{Characters} and \textit{Letters} is due
to the similarity in Rashi and Printed \textit{Characters }of some of the
letters, as depicted in Table \ref{tab:rashi-fonts}. Thus, the CNNs might
misclassify some character images as being Rashi instead of Printed and vice
versa, while relating to the same \textit{Letter}. This is considered a
\textit{Character} classification error, and a correct \textit{Letter}
classification. The proposed AutoML scheme is shown to significantly
outperform all other schemes\ while utilizing the least number of the CNN
parameters.\begin{table}[tbh]
\centering%
\begin{tabular}
[c]{|c|c|c|}\hline
\textbf{Scheme} & \textbf{Error rate} [\%] & \textbf{Number of CNN
Parameters}\\\hline\hline
\multicolumn{1}{|l|}{{\textbf{SNN}}} & 0.313 & 1.01M\\\hline
\multicolumn{1}{|l|}{\textbf{STN}} & 0.243 & 0.73M\\\hline
\multicolumn{1}{|l|}{\textbf{AlexNet}} & 0.240 & 21.82M\\\hline
\multicolumn{1}{|l|}{\textbf{AutoML}} & \textbf{0.188} & 0.72M\\\hline
\end{tabular}
\caption{The letters classification accuracy of the proposed CNNs trained
using 3M samples. STN - Spatial transform net. ALX - AlexNet. AutoML - MNIST
optimized by a Genetic Algorithm. }%
\label{table:cnn summary}%
\end{table}

\begin{figure}[tbh]
\centering
\par%
\begin{tabular}
[c]{cccccc||cccccc}%
\includegraphics[width=0.015\textwidth]{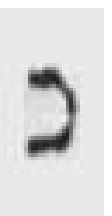} &
\includegraphics[width=0.015\textwidth]{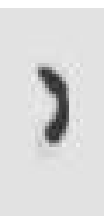} &
\includegraphics[width=0.015\textwidth]{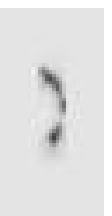} &
\includegraphics[width=0.015\textwidth]{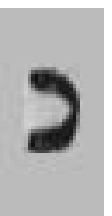} &
\includegraphics[width=0.015\textwidth]{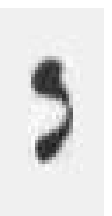} &
\includegraphics[width=0.015\textwidth]{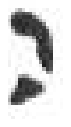} &
\includegraphics[width=0.015\textwidth]{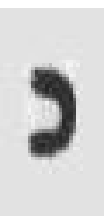} &
\includegraphics[width=0.015\textwidth]{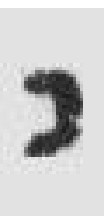} &
\includegraphics[width=0.015\textwidth]{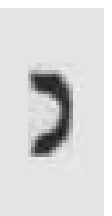} &
\includegraphics[width=0.015\textwidth]{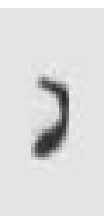} &
\includegraphics[width=0.015\textwidth]{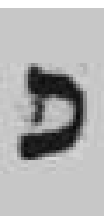} &
\includegraphics[width=0.015\textwidth]{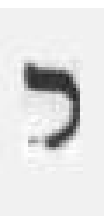}\\
\includegraphics[width=0.015\textwidth]{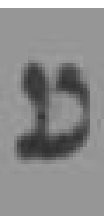} &
\includegraphics[width=0.015\textwidth]{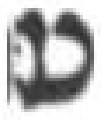} &
\includegraphics[width=0.015\textwidth]{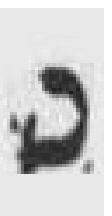} &
\includegraphics[width=0.015\textwidth]{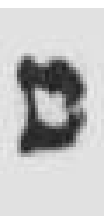} &
\includegraphics[width=0.015\textwidth]{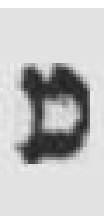} &
\includegraphics[width=0.015\textwidth]{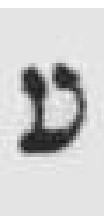} &
\includegraphics[width=0.015\textwidth]{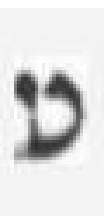} &
\includegraphics[width=0.015\textwidth]{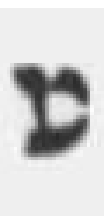} &
\includegraphics[width=0.015\textwidth]{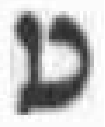} &
\includegraphics[width=0.015\textwidth]{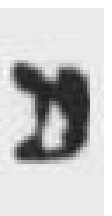} &
\includegraphics[width=0.015\textwidth]{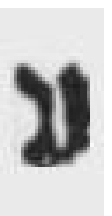} &
\includegraphics[width=0.015\textwidth]{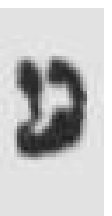}\\
\includegraphics[width=0.015\textwidth]{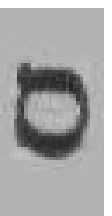} &
\includegraphics[width=0.015\textwidth]{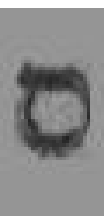} &
\includegraphics[width=0.015\textwidth]{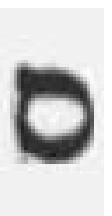} &
\includegraphics[width=0.015\textwidth]{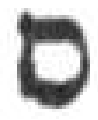} &
\includegraphics[width=0.015\textwidth]{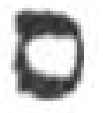} &
\includegraphics[width=0.015\textwidth]{./smk-B001-EL002_let.eps} &
\includegraphics[width=0.015\textwidth]{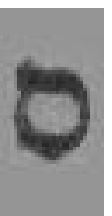} &
\includegraphics[width=0.015\textwidth]{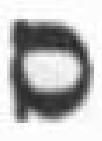} &
\includegraphics[width=0.015\textwidth]{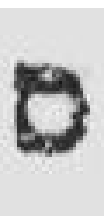} &
\includegraphics[width=0.015\textwidth]{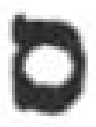} &
\includegraphics[width=0.015\textwidth]{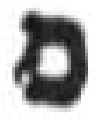} &
\includegraphics[width=0.015\textwidth]{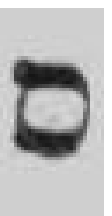}\\
\includegraphics[width=0.015\textwidth]{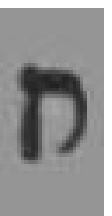} &
\includegraphics[width=0.015\textwidth]{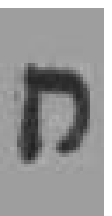} &
\includegraphics[width=0.015\textwidth]{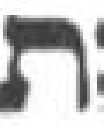} &
\includegraphics[width=0.015\textwidth]{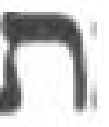} &
\includegraphics[width=0.015\textwidth]{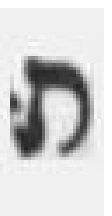} &
\includegraphics[width=0.015\textwidth]{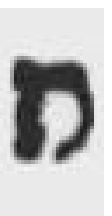} &
\includegraphics[width=0.015\textwidth]{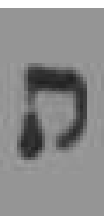} &
\includegraphics[width=0.015\textwidth]{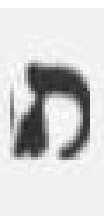} &
\includegraphics[width=0.015\textwidth]{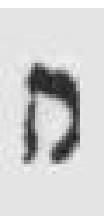} &
\includegraphics[width=0.015\textwidth]{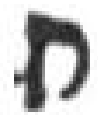} &
\includegraphics[width=0.015\textwidth]{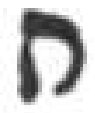} &
\includegraphics[width=0.015\textwidth]{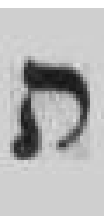}
\end{tabular}
\caption{The most misclassified characters. First row: \textcjheb{\<n|>}
('nun') vs. \textcjheb{\<k|>} ('kaf'). Second Row: \textcjheb{.t} ('tet') vs.
\textcjheb{`} ('ain') Third row: \textcjheb{s} ('smk') vs. \textcjheb{\<M>}
('mem-p'). Fourth row: \textcjheb{t} ('tav') vs. \textcjheb{.h} ('het') }%
\label{fig:difficult letters}%
\end{figure}

Table \ref{table:Candidates CNN} and Fig. \ref{fig:difficult letters} list and
depict the most misclassified Characters, and the most common
misclassifications for each Character. It follows that the misclassified
characters are indeed visually similar, and might be difficult to distinguish
even for a human observer. Hence the need\ to utilize the sequential
information using LSTM.\begin{table}[tbh]
{\footnotesize \centering%
\begin{tabular}
[c]{|p{0.05cm}|c|c|c|}\hline
\textbf{L} & \textbf{Cand \#1} & \textbf{Cand \#2} & \textbf{Cand
\#3}\\\hline\hline
\textcjheb{\<n|>} & \textcjheb{\<k|>} - 0.70\% (64\%) & \textcjheb{l} -
0.17\% (15.3\%) & \textcjheb{g} - 0.08\% (7.2\%)\\\hline
\textcjheb{\<M>} & \textcjheb{s} - 0.86\% (80\%) & \textcjheb{h} - 0.19\%
(17.3\%) & \textcjheb{.t} - 0.01\% (1.0\%)\\\hline
\textcjheb{s} & \textcjheb{\<M>} - 0.79\% (84\%) & \textcjheb{h} - 0.06\%
(6.4\%) & \textcjheb{/s} - 0.02\% (2.7\%)\\\hline
\textcjheb{.t} & \textcjheb{`} - 0.56\% (77\%) & \textcjheb{/s} - 0.05\%
(7.1\%) & \textcjheb{\<m|>} - 0.03\% (4.8\%)\\\hline
\textcjheb{.h} & \textcjheb{t} - 0.39\% (62\%) & \textcjheb{\<m|>} - 0.12\%
(18.2\%) & \textcjheb{'} - 0.08\% (12.2\%)\\\hline
\end{tabular}
}\caption{The most confused characters and their misclassifications, when
applying the AutoML CNN, without the use of LSTM or book-specific refinement.}%
\label{table:Candidates CNN}%
\end{table}

\subsection{OCR results using AutoML CNN+LSTM}

\label{subsec:LSTM Results}

Following the results of Section \ref{subsec:CNN Results}, we studied refining
the performance of the AutoML-based CNN by adding a LSTM layer to the AutoML
CNN to learn the \textit{Letters} sequences. For that we considered three
training strategies. First, we "froze" the CNN weights, (L1 in Fig.
\ref{fig:LSTM}) and trained only the LSTM layer. Denote this CNN as
AutoML-LSTM. Second, we further refined the AutoML-LSTM network, by unfreezing
its CNN weights (L1 in Fig. \ref{fig:LSTM}) and "freezing" the LSTM and FC
layers (L2 and L3) and denoted the resulting network RE-CNN. Last, we refined
the AutoML-LSTM CNN by training the entire network (L1+L2+L3 in Fig.
\ref{fig:LSTM}), and denoting the resulting network RE-BOTH.

\begin{figure}[tbh]
\centering
\includegraphics[width=0.5\textwidth]{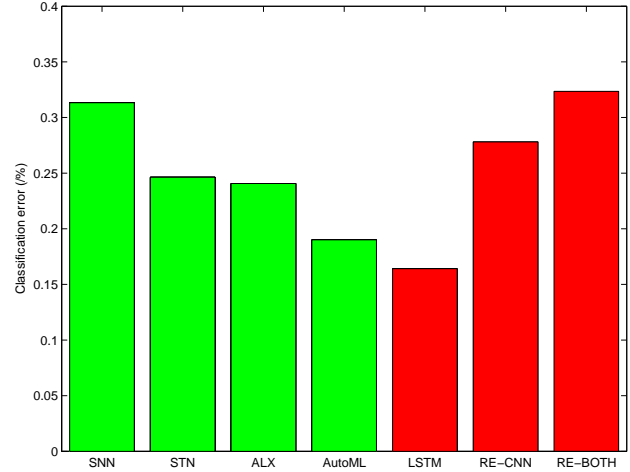}
\caption{Classification accuracy results of CNN+LSTM networks with different
training schemes. The green bars depict the accuracy of the CNN networks.
The red bars depict the further refinement of the AutoML network.
LSTM : trained by training the LSTM and FC (L2+L3 in Fig. \ref{fig:LSTM})
layers and "freezing" the CNN layers (L1 in Fig. \ref{fig:LSTM})
RE-CNN: trained by "freezing" the LSTM and FC (L2+L3 in Fig. \ref{fig:LSTM})
layers and training the CNN layers (L1 in Fig. \ref{fig:LSTM}). RE-BOTH:
training the entire network (L1+L2+L3 in Fig. \ref{fig:LSTM}).}%
\label{fig:LSTM results}%
\end{figure}

Figure \ref{fig:LSTM results} reports the classification error of the three
schemes, and compares them to the CNN-only schemes discussed in Section
\ref{subsec:CNN Results}. It follows that \ the AutoML-LSTM scheme
significantly outperforms all other training variations, and the CNN-only
schemes. 

Table \ref{table:Candidates LSTM} reports the \textit{Letters} accuracy
results using CNN+LSTM. The confusion between \textcjheb{\<n|>} and
\textcjheb{\<k|>} ('nun' and 'kaf'), was improved from 0.74\% to 0.67\% for
the CNN and LSTM, respectively.\begin{table}[tbh]
{\footnotesize \centering%
\begin{tabular}
[c]{|p{0.05cm}|c|c|c|}\hline
\textbf{L} & \textbf{Cand \#1} & \textbf{Cand \#2} & \textbf{Cand
\#3}\\\hline\hline
\textcjheb{\<n|>} & \textcjheb{\<k|>} - 0.67\% (62\%) & \textcjheb{l} -
0.17\% (15.7\%) & \textcjheb{g} - 0.07\% (6.8\%)\\
\textcjheb{s} & \textcjheb{\<M>} - 0.68\% (80\%) & \textcjheb{h} - 0.08\%
(9.4\%) & \textcjheb{/s} - 0.02\% (2.9\%)\\
\textcjheb{.t} & \textcjheb{`} - 0.54\% (76\%) & \textcjheb{/s} - 0.05\%
(7.4\%) & \textcjheb{\<m|>} - 0.03\% (4.9\%)\\
\textcjheb{\<M>} & \textcjheb{s} - 0.55\% (83\%) & \textcjheb{h} - 0.09\%
(14.4\%) & \textcjheb{/s} - 0.01\% (1.1\%)\\
\textcjheb{.h} & \textcjheb{t} - 0.36\% (60\%) & \textcjheb{\<m|>} - 0.12\%
(19.1\%) & \textcjheb{'} - 0.08\% (12.7\%)\\\hline
\end{tabular}
}\caption{The most confused letters and their erroneous detections, using the
proposed CNN + LSTM scheme.}%
\label{table:Candidates LSTM}%
\end{table}

\subsection{Book-Specific OCR Refinement}

\label{subsec:Refinement Results}

The book-specific refinement scheme was introduced in Section
\ref{subsec:Refine}, allowing to utilize the invariance of the dialect of the
Rashi scripts in contrast to the varying graphical manifestation of the
characters. For that, given a particular \textit{test} script (book), we
applied the AutoML-LSTM CNN, trained as in Section \ref{subsec:LSTM Results},
on a small subset of the \textit{test} script, and used part of these
\textit{test} results, to refine the AutoML-LSTM CNN.

Thus, for each test script we applied both the CNN layers (L1 in Fig.
\ref{fig:LSTM}) of AutoML-LSTM, and the entire AutoML-LSTM (L1+L2+L3 in Fig.
\ref{fig:LSTM}). The characters that were classified similarly by both schemes
were used as a \textit{refinement} set for the CNN layers of AutoML-LSTM.
Tables \ref{table:refinment-chars-acc} and \ref{table:refinment-letters-acc}
show the error classification rates of the proposed refinement scheme for the
\textit{Characters} and \textit{Letters}, respectively. As the refinement set
was relatively small, we repeated the experiment four times.

It follows that the proposed refinement scheme reduces the \textit{Letters}
classification error of the AutoML CNN+LSTM scheme from 0.188\% (in Table
\ref{table:cnn summary}) to 0.164 (in Table \ref{table:refinment-letters-acc}%
), exemplifying the validity of the proposed scheme. We note that the
refinement parameters corresponding to the highest \textit{Characters}
classification error (0.611\% in Tables \ref{table:refinment-chars-acc}), do
not correspond to the highest letters classification accuracy (0.164\% in
Tables \ref{table:refinment-letters-acc}). We attribute that, as in Section
\ref{subsec:CNN Results}, to the similarity of some of the \textit{Characters}
in both Rashi and Printed scripts, resulting\textit{ }in \textit{Characters}
misclassifications, but accurate \textit{Letters}
classifications.\begin{table}[tbh]
\centering
\par%
\begin{tabular}
[c]{|c||c|c|c|c|c|c|}\hline
\textbf{\#letters} & \multicolumn{6}{|c|}{\textbf{\# Refinement epochs}%
}\\\hline\hline
\multicolumn{1}{|c||}{} & \textbf{1} & \textbf{2} & \textbf{5} & \textbf{10} &
\textbf{20} & \textbf{40}\\\hline
\multicolumn{1}{|c||}{\textbf{1000}} & 0.646 & 0.669 & 0.682 & 0.699 & 0.713 &
0.732\\\hline
\multicolumn{1}{|c||}{\textbf{2000}} & 0.636 & 0.674 & 0.688 & 0.693 & 0.692 &
0.706\\\hline
\multicolumn{1}{|c||}{\textbf{5000}} & \textbf{0.611} & 0.650 & 0.677 &
0.694 & 0.691 & 0.680\\\hline
\multicolumn{1}{|c||}{\textbf{10000}} & 0.628 & 0.653 & 0.630 & 0.641 &
0.633 & 0.634\\\hline
\end{tabular}
\caption{Characters error rate percentage when applying a book-specific
refinement scheme for the AutoML-based CNN+LSTM network. the error rates are
averages over 14 books. The error rate for characters without the proposed
book-specific refinement was 0.696\%}%
\label{table:refinment-chars-acc}%
\end{table}\begin{table}[tbh]
\centering%
\begin{tabular}
[c]{|c||c|c|c|c|c|c|}\hline
\textbf{\#letters} & \multicolumn{6}{c|}{\textbf{\# Refinement epochs}%
}\\\hline\hline
& \textbf{1} & \textbf{2} & \textbf{5} & \textbf{10} & \textbf{20} &
\textbf{40}\\\hline
\textbf{1000} & 0.194 & 0.194 & 0.187 & 0.183 & 0.181 & 0.180\\\hline
\textbf{2000} & 0.187 & 0.181 & 0.179 & 0.174 & 0.174 & \textbf{0.164}\\\hline
\textbf{5000} & 0.184 & 0.176 & 0.175 & 0.171 & 0.166 & \textbf{0.164}\\\hline
\textbf{10000} & 0.184 & 0.180 & 0.180 & 0.174 & 0.175 & 0.173\\\hline
\end{tabular}
\caption{Letters error rate percentage when applying a book-specific
refinement scheme for the AutoML-based CNN+LSTM network. The error rates are
averages over 14 books. The average error rate for letters \textit{without}
LSTM\ and book-specific refinement was 0.188\%. The letters error rate without
book-specific refinement while using LSTM was 0.174\%}%
\label{table:refinment-letters-acc}%
\end{table}

Tables \ref{table:Candidates LSTM} and \ref{table:Candidates REFINE}, report
the classification errors of the most confused letters, with and without the
book-specific refinement, respectively. The classification errors of most
letters were improved, while the accuracy of the letters \textcjheb{s} and
\textcjheb{\<M>} ('smk' and 'mem-p') was not. The refinement scheme improves
the CNN phase, and the visual inference, while \textcjheb{s} and
\textcjheb{\<M>} are difficult to distinguish visually.\begin{table}[tbh]
{\footnotesize \centering%
\begin{tabular}
[c]{|c|c|c|c|}\hline
\textbf{L} & \textbf{Cand \#1} & \textbf{Cand \#2} & \textbf{Cand
\#3}\\\hline\hline
\textcjheb{\<n|>} & \textcjheb{\<k|>} - 0.42\% (49\%) & \textcjheb{l} -
0.18\% (21.3\%) & \textcjheb{g} - 0.08\% (8.9\%)\\
\textcjheb{s} & \textcjheb{\<M>} - 0.65\% (84\%) & \textcjheb{h} - 0.05\%
(6.4\%) & \textcjheb{\<p|>} - 0.02\% (2.6\%)\\
\textcjheb{.t} & \textcjheb{`} - 0.48\% (74\%) & \textcjheb{/s} - 0.05\%
(7.9\%) & \textcjheb{\<m|>} - 0.03\% (5.3\%)\\
\textcjheb{\<M>} & \textcjheb{s} - 0.71\% (89\%) & \textcjheb{h} - 0.06\%
(7.9\%) & \textcjheb{b} - 0.01\% (0.9\%)\\
\textcjheb{.h} & \textcjheb{t} - 0.32\% (63\%) & \textcjheb{'} - 0.08\%
(15.2\%) & \textcjheb{\<m|>} - 0.06\% (11.0\%)\\\hline
\end{tabular}
}\caption{The most confused letters and the corresponding erroneous
classifications of the CNN+LSTM+book-specific refinement scheme.}%
\label{table:Candidates REFINE}%
\end{table}

We show qualitative misclassification results in Table \ref{table:errors},
where some of the misclassification (examples 1, 2, 4, 5, and 14) are due to
printing errors, while others (examples 6, 7 and 12) are due to paper erosion.
Some of the misclassifications (3, 9. 10. 11 and 13 ) are quite similar
visually and difficult to distinguish even by a human
observer.\begin{table}[tbh]
\centering%
\begin{tabular}
[c]{|c|c|c|c|c|}\hline
& \textbf{True} & \textbf{False} & \textbf{Letter} & \textbf{ROI}%
\\\hline\hline
1 & \multicolumn{1}{|c|}{\textcjheb{'} (alf)} &
\multicolumn{1}{|c|}{\textcjheb{\<k|>} (kaf)} &
\includegraphics[width=0.015\textwidth]{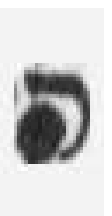} &
\includegraphics[width=0.15\textwidth]{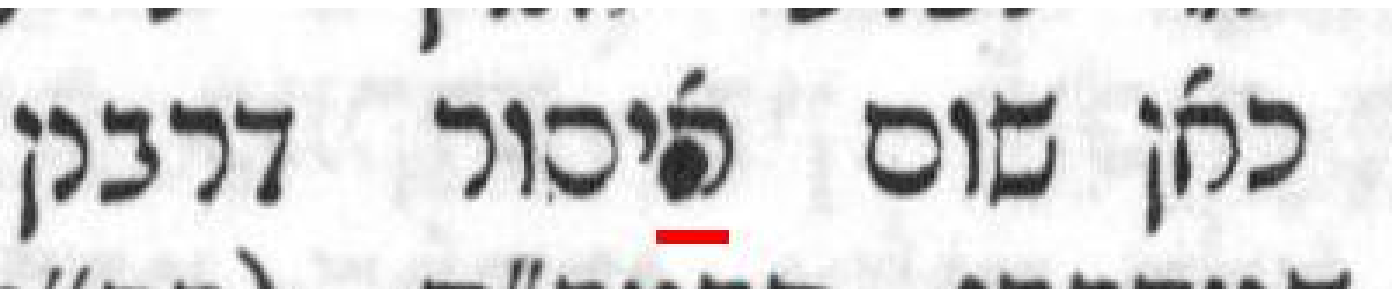}\\
2 & \multicolumn{1}{|c|}{\textcjheb{b} (bet)} &
\multicolumn{1}{|c|}{\textcjheb{'} (alf)} &
\includegraphics[width=0.015\textwidth]{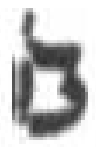} &
\includegraphics[width=0.15\textwidth]{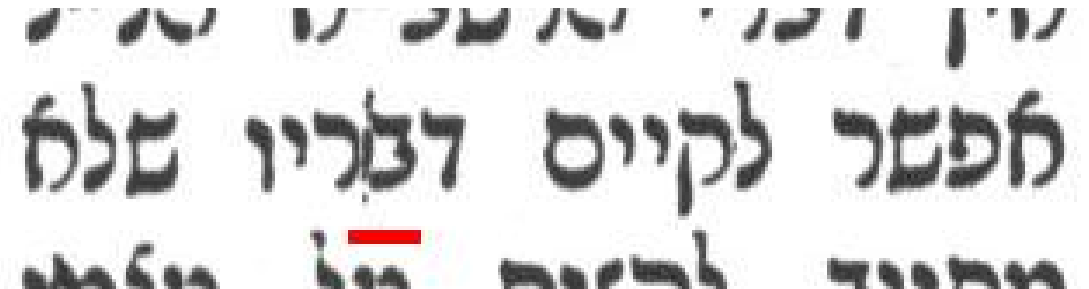}\\
3 & \multicolumn{1}{|c|}{\textcjheb{b} (bet)} &
\multicolumn{1}{|c|}{\textcjheb{g} (gml)} &
\includegraphics[width=0.015\textwidth]{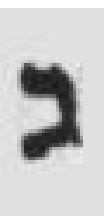} &
\includegraphics[width=0.15\textwidth]{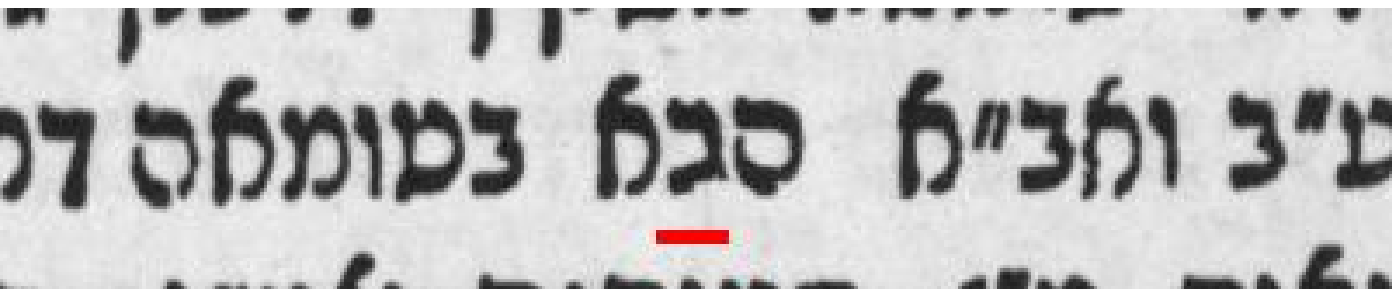}\\
4 & \multicolumn{1}{|c|}{\textcjheb{b} (bet)} &
\multicolumn{1}{|c|}{\textcjheb{\<k|>} (kaf)} &
\includegraphics[width=0.015\textwidth]{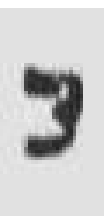} &
\includegraphics[width=0.15\textwidth]{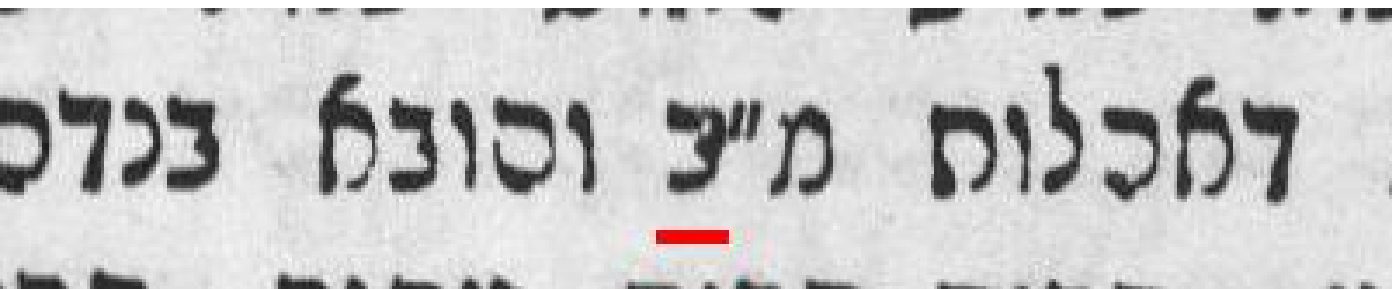}\\
5 & \multicolumn{1}{|c|}{\textcjheb{b} (bet)} &
\multicolumn{1}{|c|}{\textcjheb{\<M>} (mem-p)} &
\includegraphics[width=0.015\textwidth]{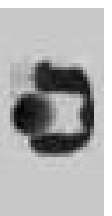} &
\includegraphics[width=0.15\textwidth]{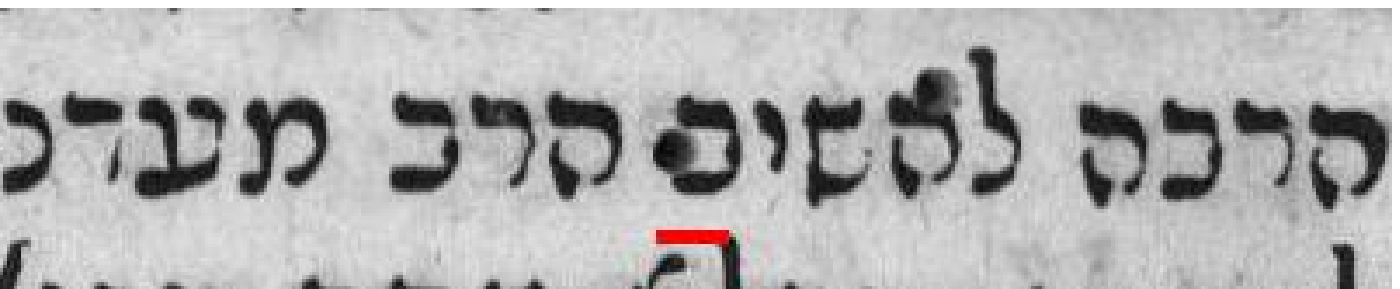}\\
6 & \multicolumn{1}{|c|}{\textcjheb{g} (gml)} &
\multicolumn{1}{|c|}{\textcjheb{\<n|>} (nun)} &
\includegraphics[width=0.015\textwidth]{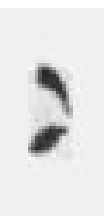} &
\includegraphics[width=0.15\textwidth]{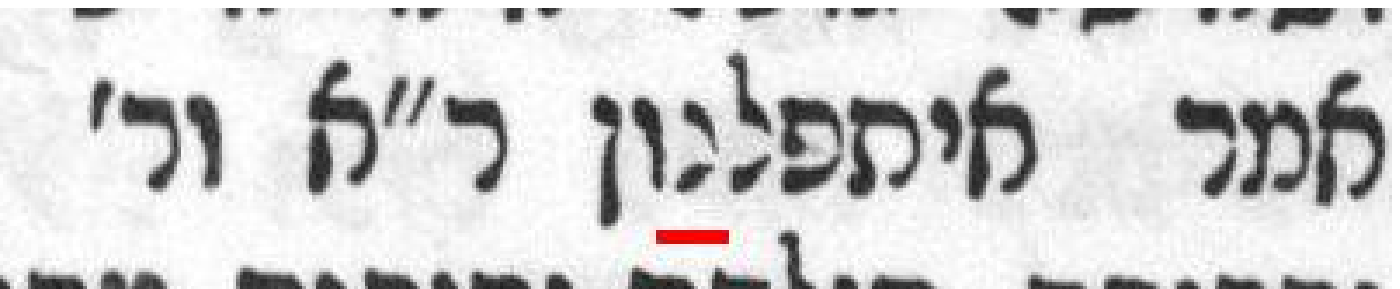}\\
7 & \multicolumn{1}{|c|}{\textcjheb{d} (dlt)} &
\multicolumn{1}{|c|}{\textcjheb{w} (vav)} &
\includegraphics[width=0.015\textwidth]{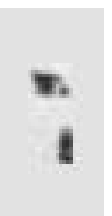} &
\includegraphics[width=0.15\textwidth]{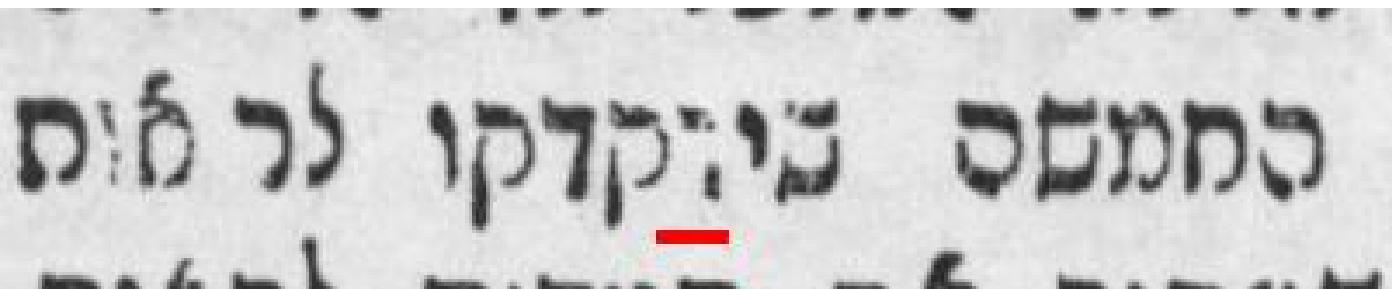}\\
8 & \multicolumn{1}{|c|}{\textcjheb{h} (hea)} &
\multicolumn{1}{|c|}{\textcjheb{b} (bet)} &
\includegraphics[width=0.015\textwidth]{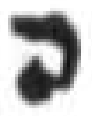} &
\includegraphics[width=0.15\textwidth]{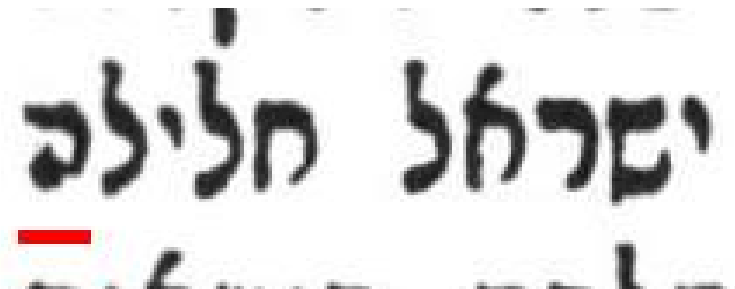}\\
9 & \multicolumn{1}{|c|}{\textcjheb{.h} (het)} &
\multicolumn{1}{|c|}{\textcjheb{t} (tav)} &
\includegraphics[width=0.015\textwidth]{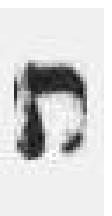} &
\includegraphics[width=0.15\textwidth]{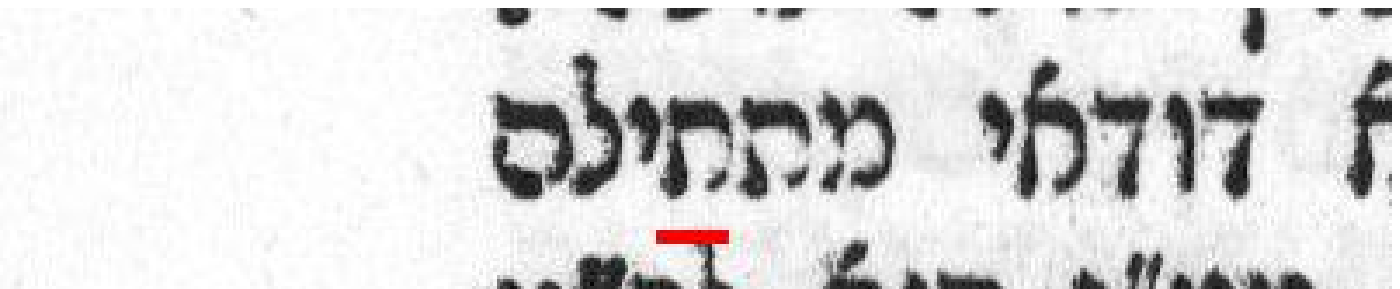}\\
10 & \multicolumn{1}{|c|}{\textcjheb{.t} (tet)} &
\multicolumn{1}{|c|}{\textcjheb{`} (ain)} &
\includegraphics[width=0.015\textwidth]{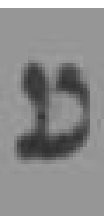} &
\includegraphics[width=0.15\textwidth]{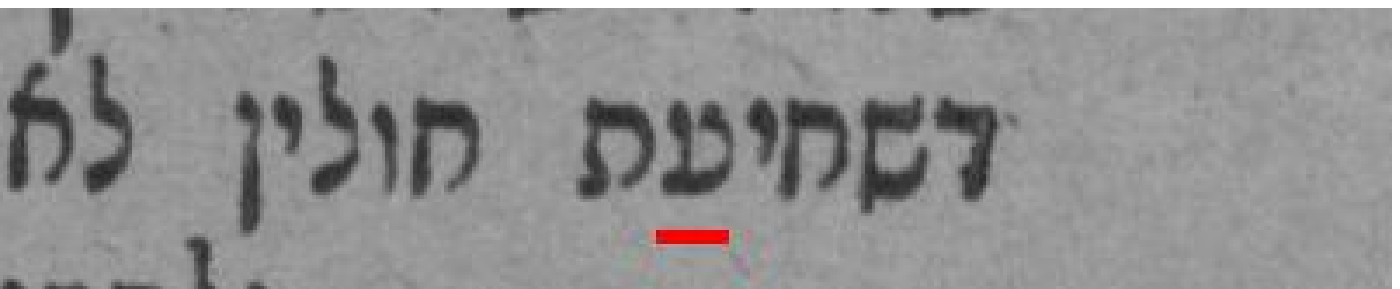}\\
11 & \multicolumn{1}{|c|}{\textcjheb{.t} (tet)} &
\multicolumn{1}{|c|}{\textcjheb{`} (ain)} &
\includegraphics[width=0.015\textwidth]{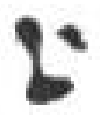} &
\includegraphics[width=0.15\textwidth]{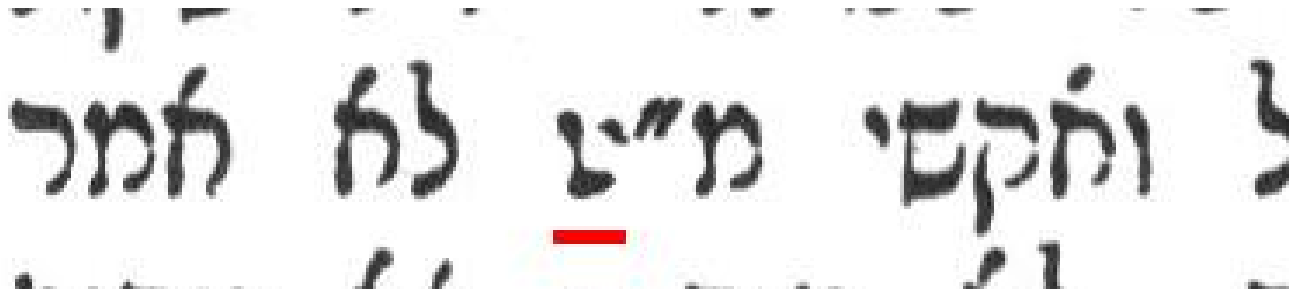}\\
12 & \multicolumn{1}{|c|}{\textcjheb{l} (lmd)} &
\multicolumn{1}{|c|}{\textcjheb{y} (yod)} &
\includegraphics[width=0.015\textwidth]{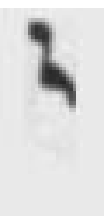} &
\includegraphics[width=0.15\textwidth]{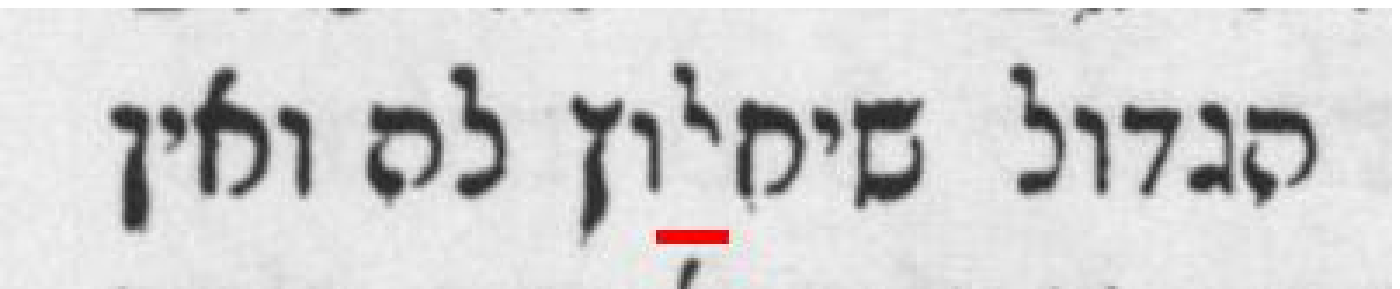}\\
13 & \multicolumn{1}{|c|}{\textcjheb{\<m|>} (mem)} &
\multicolumn{1}{|c|}{\textcjheb{`} (ain)} &
\includegraphics[width=0.015\textwidth]{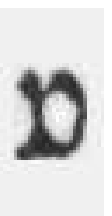} &
\includegraphics[width=0.15\textwidth]{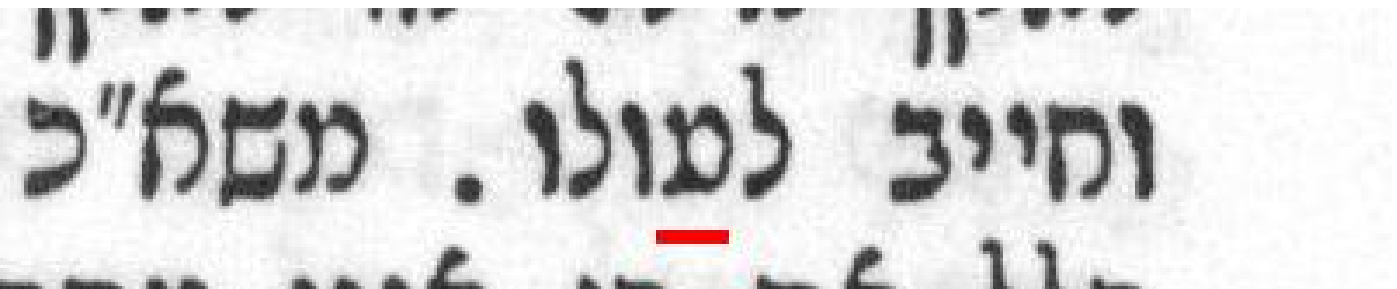}\\
14 & \multicolumn{1}{|c|}{\textcjheb{\<N>} (nun-p)} &
\multicolumn{1}{|c|}{\textcjheb{\<n|>} (nun)} &
\includegraphics[width=0.015\textwidth]{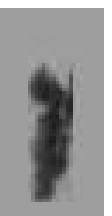} &
\includegraphics[width=0.15\textwidth]{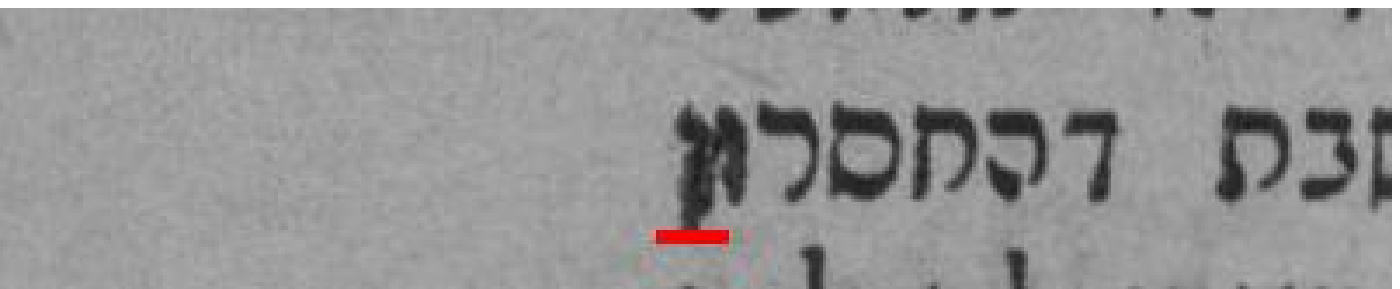}\\\hline
\end{tabular}
\caption{Representative examples of the misclassifications, of the
CNN+LSTM+book-specific refinement scheme. The symbol in question is shown in
the third column, and the corresponding ROI extracted from the manuscript is
shown in the right column.}%
\label{table:errors}%
\end{table}

\subsection{Implementation issues}

The CNN  (L1 in Fig.  \ref{fig:LSTM}) and the LSTM  (L2 + L3 in Fig.  \ref{fig:LSTM}) layers of the proposed scheme
were trained and implemented using the Keras wrapper \cite{chollet2015keras} for
TensorFlow \cite{tensorflow2015-whitepaper}.
The proposed AutoML refinement scheme introduced in Section \ref{subsec:automl},
that is based on a Genetic Algorithm, was implemented in Matlab and
MatConvNet \cite{vedaldi15matconvnet}. All of the training was conducted on Titan X Maxwell and the
training of the AutoML CNN required four hours.

\section{Conclusions}

\label{sec:conclusions}

In this work we proposed a Deep Learning based approach for the OCR of
manuscripts printed in the Rashi and Printed Hebrew scripts. For that, we
studied and compared four CNN architectures. In particular, we learn both the
characters\ visual appearance and the (apriori unknown) vocabulary using CNNs
and LSTM, respectively. We derive an AutoML scheme based on Genetic Algorithm
to optimize the CNN architecture with respect to the classification accuracy.
The resulting CNN is shown to compare favourably with other CNNs based on
larger (AlexNet) or similar nets. We also propose a book-specific approach for
refining the OCR CNN per book. The resulting scheme is shown to significantly
outperform similar unoptimized CNNs and achieves OCR accuracy of 99.840\%.

OCR is a particular example of the Structured Image Classification problem
that has multiple applications in computer vision
\cite{DBLP:journals/corr/GoldmanG17} and medical imaging \cite{zebra}. In
future we aim to further develop and apply the proposed AutoML and
book-specific refinement schemes in that context.

\bigskip\FloatBarrier
\bibliographystyle{ieeetr}
\bibliography{AUTOML_RASHI_OCR}

\end{document}